\documentclass[letterpaper]{article}
\usepackage{aaai}
\usepackage{times}
\usepackage{helvet}
\usepackage{courier}
\usepackage{array}
\usepackage{tabularx}
\usepackage{caption}
\usepackage{comment}
\usepackage{verbatim}
\usepackage{booktabs}
\usepackage{subcaption} 
\usepackage{graphicx}
\usepackage{amsmath}
\usepackage{amsthm}
\usepackage{booktabs}
\usepackage{algorithm}
\usepackage{algorithmic}
\usepackage[switch]{lineno}
\usepackage{float}
\usepackage{url}
\usepackage{color}
\usepackage{tcolorbox}
\usepackage[utf8]{inputenc}

\begin{document}
%
\title{OmniAcc: Personalized Accessibility Assistant Using Generative AI \vspace{-20pt}}
\author{
Siddhant	Karki,
Ethan	Han,
Nadim	Mahmud,
Suman	Bhunia,
John	Femiani,
Vaskar	Raychoudhury\\
Miami University of Ohio, Oxford, OH 45056, USA\\
\url{karkiss@miamioh.edu,
ethanhan2003@gmail.com,
mahmudm2@miamioh.edu}\\
\url{bhunias@miamioh.edu,
femianjc@miamioh.edu,
raychov@MiamiOH.edu}
}


\maketitle
\vspace{-30pt}

\begin{abstract}
\begin{quote}
Individuals with ambulatory disabilities often encounter significant barriers when navigating urban environments due to the lack of accessible information and tools.  This paper presents OmniAcc, an AI-powered interactive navigation system that utilizes GPT-4, satellite imagery, and OpenStreetMap data to identify, classify, and map wheelchair-accessible features such as ramps and crosswalks in the built environment. OmniAcc offers personalized route planning, real-time hands-free navigation, and instant query responses regarding physical accessibility. By using zero-shot learning and customized prompts, the system ensures precise detection of accessibility features, while supporting validation through structured workflows. This paper introduces OmniAcc and explores its potential to assist urban planners and mobility-aid users, demonstrated through a case study on crosswalk detection. With a crosswalk detection accuracy of 97.5\%, OmniAcc highlights the transformative potential of AI in improving navigation and fostering more inclusive urban spaces.

\end{quote}
\end{abstract}

\section{Introduction}
Traveling across the built environment, especially in unfamiliar surroundings, is challenging for wheelchair users because of unknown barriers~\cite{ding2007design,volkel2008mobility} posed by the path as well as the destinations like public buildings. Uneven surfaces such as roads and sidewalks, the presence of stairs and steep slopes, the absence of curb cuts and pedestrian crosswalks, and varied weather conditions thwart successful outings for someone using a wheelchair. Similarly, accessible parking locations and accessible entrance(s) to a building or other public places are not always readily evident, and wheelchair users often need to look around for pointers or ask people, and sometimes there is no one to ask~\cite{hammel2015environmental}. 

\begin{figure*}[htb]
    \centering
    \includegraphics[width=0.7\linewidth]{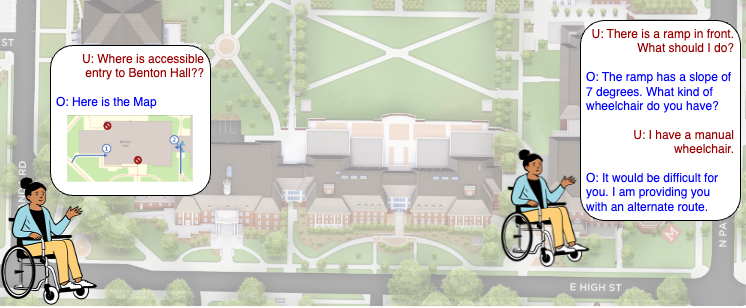}
    \caption{The overarching goal of OmniAcc. Example of a conversation between a wheelchair user and the OmniAcc Chatbot. A key enabler of this is the automatic identification of features such as crosswalks. }
    \label{fig:motiv-example}
\end{figure*}

Our research team performed I-CORP interviews with 53 participants \footnote{\url{https://uwm.edu/lubar-entrepreneurship-center/deep-dive-mypath/}} including wheelchair users, their family members, accessibility advocates, and related experts who have repeatedly pointed out that the challenges of accessible end-to-end navigation persist despite the increasing focus on built environment accessibility. A college student using a mobility scooter said - “\textit{I’m majoring in Economics…I was originally a Physics major…I wanted to be a Physicist, or an engineer, but the building in which all the Physics classes and labs were, was completely inaccessible and all my creativity could not…fix that and so, unfortunately, I was forced to change majors because of that}”.
Due to the absence of sufficient credible prior information available on any common mapping platform, mobility-aid users often need to plan extensively before an outing. They usually browse through maps, street views, publicly available user reviews, forum discussions, and building plans, and/or call the destination facility to inquire about the current accessibility status of a destination facility. In addition, our I-CORP interview participants also felt the need for a navigation system that facilitates human-like query-response interaction in real-time with personalized routing support. This is required to help users with unforeseen routing challenges due to sudden temporary barriers such as roadblocks and diversions, etc. See Figure~\ref{fig:motiv-example} for a use-case scenario. 

In this paper, we introduce the novel OmniAcc system built to provide AI-based real-time and interactive navigational assistance primarily to wheeled mobility-aid users. However, such a system requires reliable and extensive accessibility information. We plan to gradually enhance the OmniAcc system with accessibility information tapped from various reliable sources while focusing on saturating small geographical areas. OpenStreetMap (OSM) is a primary data source with user-contributed accessibility features or barriers, such as surface types, inclines, accessible entrances, stairs, ramps, crosswalks with/without pedestrian signals, etc. However, our close inspection reveals that OSM data is often incorrect, outdated, or just incomplete. Manually investigating and incorporating or updating all these path features in the OSM is time-consuming and infeasible for large geographic areas. Scalable detection and update of path features into the OSM are required for the successful operation of the OmniAcc system. In this paper, we specifically consider crosswalk as a path feature and focus on its successful detection from geospatial imagery (high-resolution GeoTIFF images). Once a crosswalk is detected correctly, it will be entered in the OSM and used for the route generation algorithm. 


Researchers have addressed the problem of crosswalk detection from satellite imagery. Verma et al. ~\cite{verma2024crosswalk} used deep learning, specifically the YOLO v5 object detection model, to automate crosswalk detection from satellite imagery. Antwi et al.~\cite{antwi2024automated} developed an automated GIS-based framework using bi-temporal aerial imagery and a YOLO-based model to detect and update crosswalk changes in Florida counties. Hosseini et al.~\cite{hosseini2023mapping} introduced TILE2NET, an open-source tool that uses semantic segmentation to extract sidewalk, crosswalk, and footpath data from high-resolution aerial imagery, enabling scalable and cost-effective pedestrian network generation for urban planning.
In addition, Ning et al.~\cite{ning2022sidewalk} developed a novel method combining aerial and street view imagery using convolutional neural networks to extract and refine sidewalk networks. Moran~\cite{moran2022crosswalk} utilized satellite imagery to map crosswalk coverage across San Francisco, revealing uneven spatial distribution and disparities between neighborhoods. While these studies effectively detect crosswalks, most rely on large training datasets, which are not always reliable or readily available, limiting their scalability and applicability.

In contrast to the above, OmniAcc incorporates OpenAI's GPT-4o model \cite{islam2024gpt} as a general purpose multimodal model that allows zero-shot crosswalk detection, with labeled data used exclusively for evaluation. 
While LLMs have proven themselves as powerful low shot and zero shot classifiers, 
to the best of our knowledge, this is the first such attempt at applying multimodal large language models specifically for zero-shot \emph{road} feature detection.  
In particular, we incorporate \emph{visual} prompting strategies by exploring different overlays or annotations on the imagery in addition to traditional text-based prompt engineering. 
We chose the GPT-4o model due to its advanced multimodal understanding, which makes it well-suited for comprehending textual and image inputs for detecting wheelchair-accessible features. The use of high resolution orthophoto imagery is also useful for our solution, as it provides precise geospatial metadata and high-resolution detail necessary for identifying accessible features, such as crosswalks. 
We evaluate crosswalk detection on data collected from Oxford, Ohio, showcasing excellent performance with an accuracy of 97.5\%.



In this paper, we make the following unique contributions.

\begin{itemize}
    \item We propose OmniAcc, a novel Gen-AI-based real-time and interactive routing and navigation system for wheeled mobility-aid users. 
    \item We demonstrate that effective prompting strategies in a multimodal large language model (OpenAI’s GPT-4o model) enable zero-shot learning. We do this by efficiently processing high-resolution aerial imagery to identify and classify crosswalks without learning from labeled data. 

    \item We explore and evaluate novel visual prompting strategies, which go beyond text based prompt engineering by incorporating visual overlays to determine which annotations on the imagery are most effective in guiding the zero shot task. 
    
    \item Our extensive experimentation with different images shows that OmniAcc system can detect crosswalks with 97.5\% accuracy, this is a very significant result in zero-shot image recognition. 
\end{itemize}

\section{Related Work}

Research on enhancing urban accessibility focuses on three main areas: (1) pedestrian infrastructure detection, (2) generative AI and few-shot or zero-shot learning, and (3) multimodal AI models. These approaches utilize advanced AI techniques to improve urban mobility and accessibility for individuals with disabilities.

\subsection{Pedestrian Infrastructure Detection from Aerial Imagery}

Several studies have explored using satellite and aerial imagery, along with computer vision, to detect and map pedestrian infrastructure like crosswalks, sidewalks, and footpaths. Verma et al. ~\cite{verma2024crosswalk} used YOLO v5 for crosswalk detection, achieving accuracies of 71\% in Washington, D.C., and 89\% in Los Angeles. Antwi et al. ~\cite{antwi2024automated} developed a GIS-based framework for detecting crosswalk changes using YOLOv2, identifying over 2,000 crosswalk changes in Florida. Hosseini et al. ~\cite{hosseini2023mapping} created TILE2NET, a tool for generating pedestrian infrastructure datasets from aerial imagery with an average 84.51\% IoU accuracy. Ning et al. ~\cite{ning2022sidewalk} combined aerial and street view imagery to improve sidewalk detection, linking 20\% more disconnected segments. Moran et al. ~\cite{moran2022crosswalk} assessed crosswalk distribution in San Francisco, highlighting inequities in coverage across neighborhoods.

\subsection{Generative AI and Few-Shot Learning in Remote Sensing}

Generative AI models, especially large language models (LLMs), have been used for remote sensing applications to reduce reliance on labeled data. Islam et al. ~\cite{islam2024gpt} demonstrated the capabilities of GPT-4o in various domains, including education and healthcare, due to its ability to process multimodal data efficiently. Qiu et al.~\cite{qiu2024few} explored few-shot learning for remote sensing scene classification, showing the potential of vision-language models (VLMs) like BLIP and CLIP for improving classification accuracy with minimal labeled data. Their research highlights the promise of VLMs in addressing challenges in remote sensing classification tasks.

\subsection{Zero-shot Learning for Vision-Language Models}

Recent advances in zero-shot learning have enabled vision-language models (VLMs) to excel in specialized tasks without task-specific fine-tuning. GeoChat ~\cite{kuckreja2024geochat}, MMCPF  ~\cite{tang2024chain}, and ALFA ~\cite{zhu2024llms} exemplify this by addressing distinct challenges in remote sensing, camouflaged object detection, and anomaly detection, respectively. GeoChat uses a unique dataset of 318k image-text pairs and adapts the LLaVA-1.5 framework to achieve strong multitasking conversational capabilities, excelling in spatial reasoning and region-specific tasks. Multimodal Camo-Perceptive Framework (MMCPF) employs Multimodal Large Language Models (MLLMs) and introduces the Chain of Visual Perception (CoVP) to enhance camouflaged object detection using linguistic prompts and visual completion, surpassing state-of-the-art methods. Meanwhile, ALFA utilizes Large Vision-Language Models (LVLMs) with run-time prompt adaptation and fine-grained alignment to achieve precise anomaly detection on challenging datasets like MVTec and VisA.

While all three approaches utilize pre-trained foundation models, their innovations in prompt design, alignment mechanisms, and dataset adaptation underscore their versatility. GeoChat sets a benchmark for multitasking in remote sensing, MMCPF demonstrates the power of multimodal cues in complex visual tasks, and ALFA highlights semantic precision in anomaly detection. Together, they showcase the potential of zero-shot learning across diverse and complex domains.

\subsection{Multimodal AI Models for Remote Sensing and Scene Understanding}

Multimodal AI models combine visual and textual data for improved scene understanding. Osco et al. ~\cite{osco2023potential} evaluated Visual ChatGPT for remote sensing tasks, revealing its moderate success but the need for domain-specific optimization. Alayrac et al. ~\cite{alayrac2022flamingo} introduced Flamingo, a VLM that outperformed existing models in few-shot tasks, offering an efficient approach for multimodal tasks with fewer annotations. De Curtò et al. ~\cite{de2023semantic} developed a system for real-time semantic scene understanding using LLMs and VLMs, showcasing the potential for generating detailed scene descriptions from UAV-captured imagery.

OmniAcc distinguishes itself from these works by using OpenAI's GPT-4o for zero-shot learning and multimodal processing, allowing for high-accuracy accessibility feature detection (e.g., crosswalks) from minimal labeled data, unlike traditional models like YOLOv5, which require large annotated datasets. While studies like those by Verma et al. (2024) and Antwi et al. (2024) focus on infrastructure detection, OmniAcc is a real-time navigation system that provides dynamic, interactive route planning tailored to mobility-aid users. It combines GeoTIFF imagery and OpenStreetMap (OSM) data for on-the-fly detection and updates, offering a more user-centric, real-time experience compared to static data used in related works.

\section{Methodology for Path Feature Detection}

In this section, we outline the operational methodology of zero-shot labeling for OmniAcc, detailing each step of the process, including data collection, preprocessing, and prompt engineering, to demonstrate how the system efficiently classifies wheelchair-accessible features in satellite imagery.

\subsection{Overview}

Our approach combines high-resolution satellite imagery,  data processing workflows, and a multimodal language-vision model (OpenAI's GPT-4o) to accurately classify wheelchair-accessible features such as crosswalks. The system integrates geospatial data, generative AI, and prompt engineering techniques to optimize performance and ensure reliable classifications.

The main dataset for OmniAcc is derived from high-resolution one-meter-per-pixel GeoTIFF images sourced from the USGS EarthExplorer NAIP dataset \cite{bhatt2023comparison}. These images are used with OpenStreetMap(OSM) to extract key urban features such as intersections where crosswalks are common.
Using OSM, road data is vectorized into graph representations where nodes represent intersections, and edges represent roads or pathways \cite{audebert2017joint}. This structured data allows for precise image segmentation of crosswalks, dividing the original GeoTIFF images into several, smaller patches that are then organized into two classes: ``crosswalk" (if a crosswalk is present) and ``not-crosswalk," with a balanced number of examples from each class in order to evaluate the model.

\begin{figure}[t]
    \centering
    \includegraphics[width=\columnwidth]{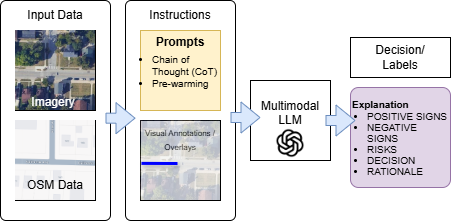}
    \caption{Overview of zero-shot labeling using OmniAcc}
    \label{fig:zslabeling}
\end{figure}

For this study, we chose crosswalks as an example because OSM has some labeled examples (although some regions are missing or inaccurately labeled).  Since some labels exist, a low-shot machine learning approach is possible, but we use a zero-shot approach because our goal is to establish a process that will work for novel features such as tactile paving, access ramps, stairs, potholes, broken sidewalks, etc. in future work. 

A key aim of our work is to assess if a foundational multimodal language-vision model (OpenAI's GPT-4o) is able to identify features in areal imagery that relate to accessibility. Figure~\ref{fig:zslabeling} outlines the overview of the zero-shot technique adopted by OmniAcc to label crosswalks. We use a chain of thought (CoT) technique to describe the specific features of interest we would like the LLM to look for in order to identify the presence of crosswalks, thus giving some guidance on what patterns to focus on. The model processes image patches encoded in base64 format, paired with tailored textual prompts. A system prompt provides an overarching description of the task.  The model then generates outputs that include binary classifications (e.g., ``CROSSWALK: Yes") along with detailed rationales that explain the model’s decision-making process. These rationales highlight visual features such as parallel white rectangles for crosswalks, as well as potential issues like shadows that may obscure these features. These outputs are validated against ground truth labels, ensuring that the classification results are accurate and interpretable. An example of the structured reasoning process are shown in Figure~\ref{fig:gpt4o_rationale_overlaid}. 

Our system prompt describes visual annotations that we overlay on the images to indicate spatial constraints (such as a "blue line" ending at a "red dot"), ensuring the analysis remains focused and contextually grounded. The prompt provides step-by-step instructions, systematically leading the model through criteria definition, observation, error checking, and final decision-making. Error mitigation is emphasized by highlighting potential confounding factors, such as shadows, lane markings, and road wear, to ensure the model accounts for ambiguity and avoids misclassification. Additionally, the inclusion of positive and negative evidence, combined with examples of structured reasoning, encourages balanced evaluation and interpretability in the model's outputs. 

OmniAcc’s performance is evaluated using standard metrics, including precision, recall, F1-score, and accuracy. Misclassifications are visualized through confusion matrices, and their causes are further investigated through the model's rationales and the corresponding visual inputs. This thorough analysis helps identify areas for improvement, such as misinterpretation of environmental features or ambiguous patterns.

By leveraging this comprehensive methodology, OmniAcc aims to provide accurate and scalable classification of wheelchair-accessible features in satellite imagery, contributing to improved urban planning and navigation for all users.

\subsection{Data Collection \& Image Patch Generation}
\begin{figure}[t]
    \centering
    \includegraphics[width=\columnwidth]{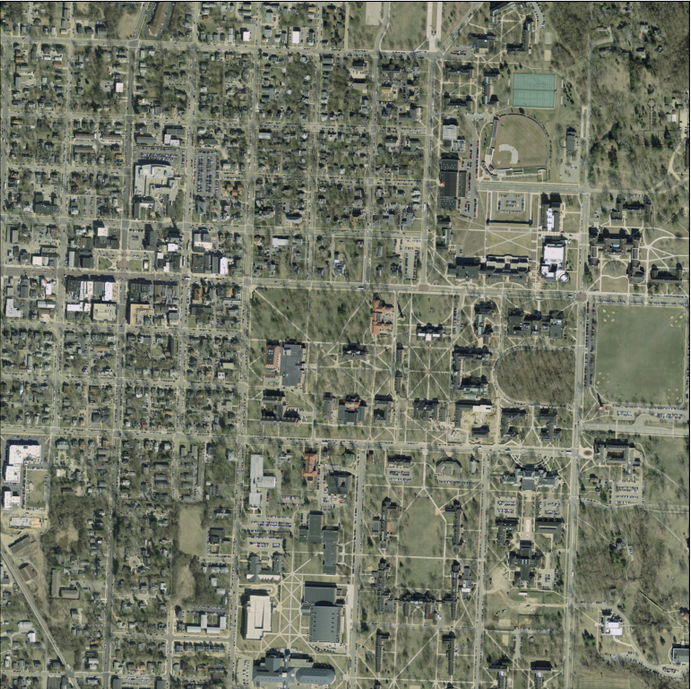}
    \caption{Original image of our study site, Oxford OH, extracted from USGS}
    \label{fig:sliding_window}
\end{figure}
The raw satellite imagery used in our study was obtained from the United States Geological Survey (USGS), offering high-resolution (5000x5000 pixels) images of both urban and rural areas in Oxford. Geospatial data, including road networks and crosswalks, were sourced from OpenStreetMap (OSM), a comprehensive and widely recognized open geospatial database. By combining these two datasets, we created several variations to test OmniAcc, employing different preprocessing techniques.

\begin{figure*}[htb]
    \centering
    \begin{subfigure}[b]{0.3\textwidth}
        \centering
        \includegraphics[width=\textwidth]{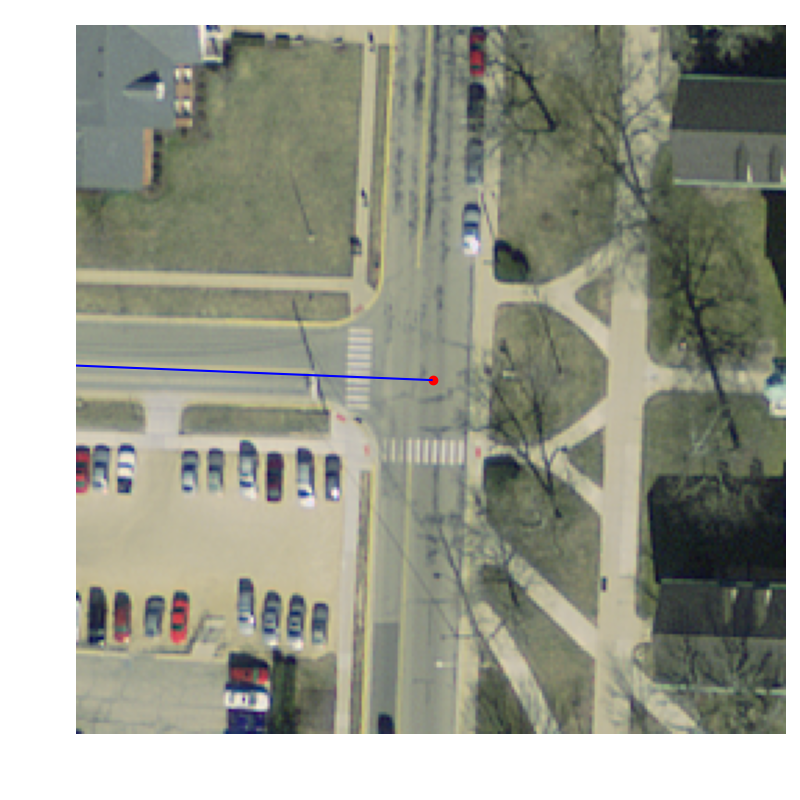}
        \caption{Patch with road orientation 1.}
        \label{fig:orientation1}
    \end{subfigure}
    \hfill
    \begin{subfigure}[b]{0.3\textwidth}
        \centering
        \includegraphics[width=\textwidth]{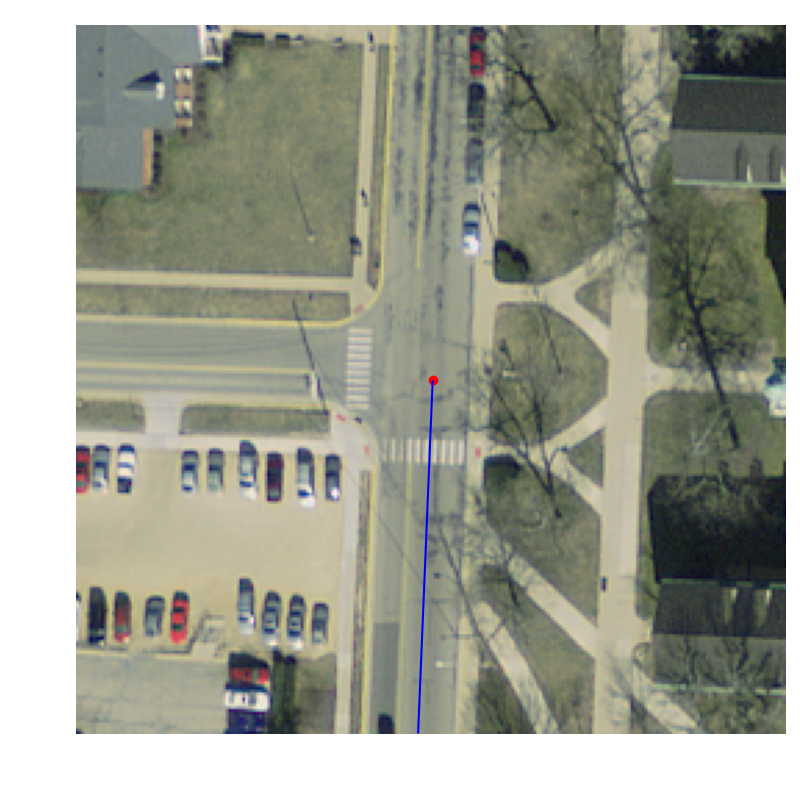}
        \caption{Patch with road orientation 2.}
        \label{fig:orientation2}
    \end{subfigure}
    \hfill
    \begin{subfigure}[b]{0.3\textwidth}
        \centering
        \includegraphics[width=\textwidth]{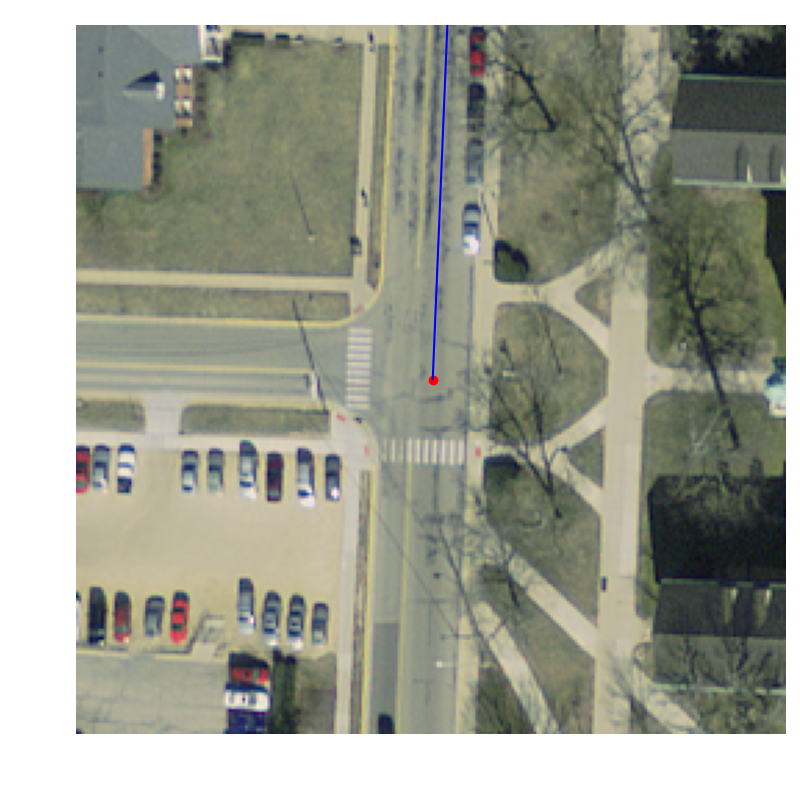}
        \caption{Patch with road orientation 3.}
        \label{fig:no_orientation}
    \end{subfigure}
    \caption{Examples of crosswalk patches treated as separate testing examples for each road direction}
    \label{fig:road_orientations}
\end{figure*}

To manage the large image size, we divided the original 5000x5000 pixel image into smaller 256x256 pixel patches. These patches were extracted by sliding a window across the larger image in both horizontal and vertical directions. In areas where multiple road orientations appeared within a patch, each orientation was treated as a separate training sample (as illustrated in Figure~\ref{fig:road_orientations}). This strategy ensured that the model could effectively learn to detect crosswalks in various directional contexts. Additionally, each patch underwent specific preprocessing steps to generate different versions of the dataset, enabling us to assess the impact of these variations on the model’s performance.

\subsection{Data Pre-processing}

To assess the influence of preprocessing on model performance, we created four distinct dataset variations, each exploring a different preprocessing approach:

\begin{itemize}
    \item Plain Dataset: This is the baseline dataset, consisting of raw satellite imagery without any modifications or additional data. It serves as a control to measure the model's performance with unprocessed, natural images.

    \item Separated Dataset: In this variation, the satellite images (raster data) and the road network overlays (vector data) from OpenStreetMap (OSM) were separated into distinct datasets. This setup enabled us to evaluate whether the model could independently learn patterns from each modality and effectively combine them during the evaluation phase.

    \item Overlaid Dataset: For this dataset, road network overlays from OSM were superimposed directly onto the satellite images. The overlays provided essential directional context and helped the model focus on spatial relationships and orientation-based features critical for crosswalk detection. To ensure the accuracy of this dataset, each image was manually validated and classified as either containing a crosswalk or not, addressing potential inconsistencies due to the crowd-sourced nature of OSM data.

    \item Blurred Dataset: In this variation, non-essential regions of the images were blurred to highlight the roads and crosswalks while reducing background noise. A Gaussian blur ($\sigma = 5$) was applied to the irrelevant areas, simplifying the visual input and guiding the model’s attention toward the most relevant features. This preprocessing step was particularly effective in improving detection accuracy by minimizing distractions.

    \begin{figure*}[htb] 
    \centering
    \begin{subfigure}[t]{0.24\linewidth} 
        \centering
        \fbox{\includegraphics[width=\linewidth]{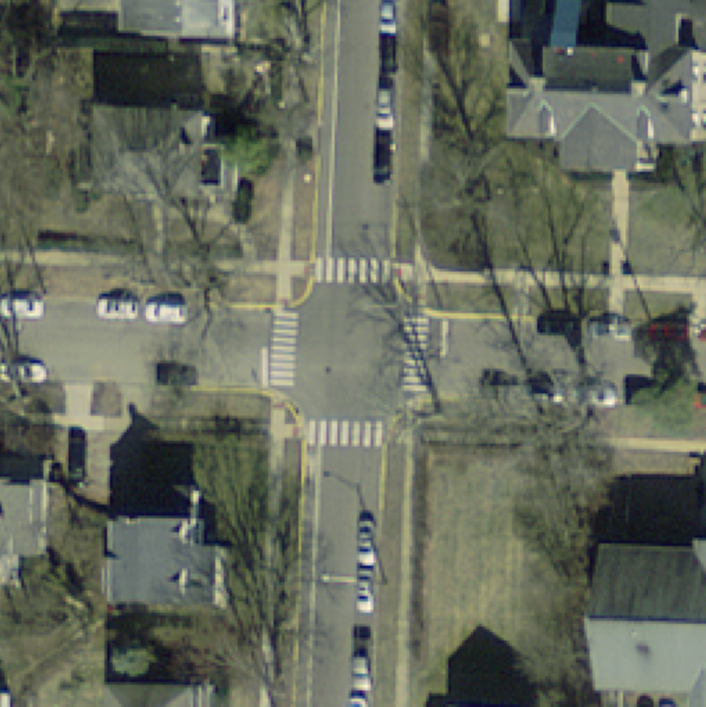}}
        \caption{Plain Image (Raster): The raster image component from the Plain Dataset.}
        \label{fig:separated_raster_example}
    \end{subfigure}
    \hfill
    \begin{subfigure}[t]{0.24\linewidth} 
        \centering
        \fbox{\includegraphics[width=\linewidth]{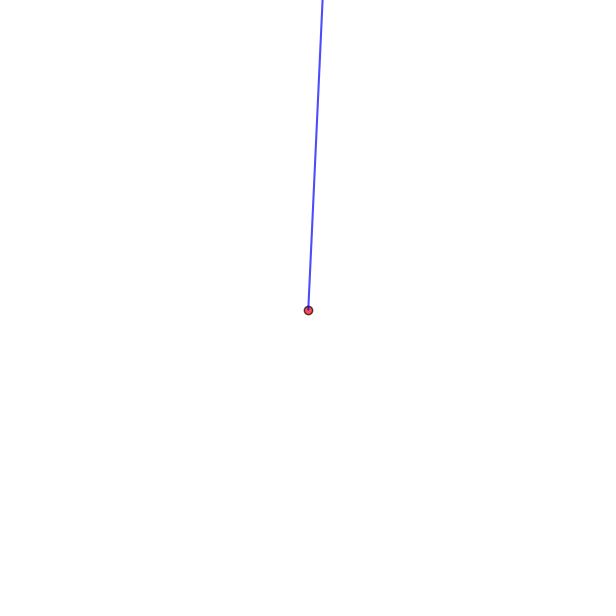}}
        \caption{Separated Dataset (Vector): The vector overlay component from the Separated Dataset.}
        \label{fig:separated_vector_example}
    \end{subfigure}
    \hfill
    \begin{subfigure}[t]{0.24\linewidth} 
        \centering
        \fbox{\includegraphics[width=\linewidth]{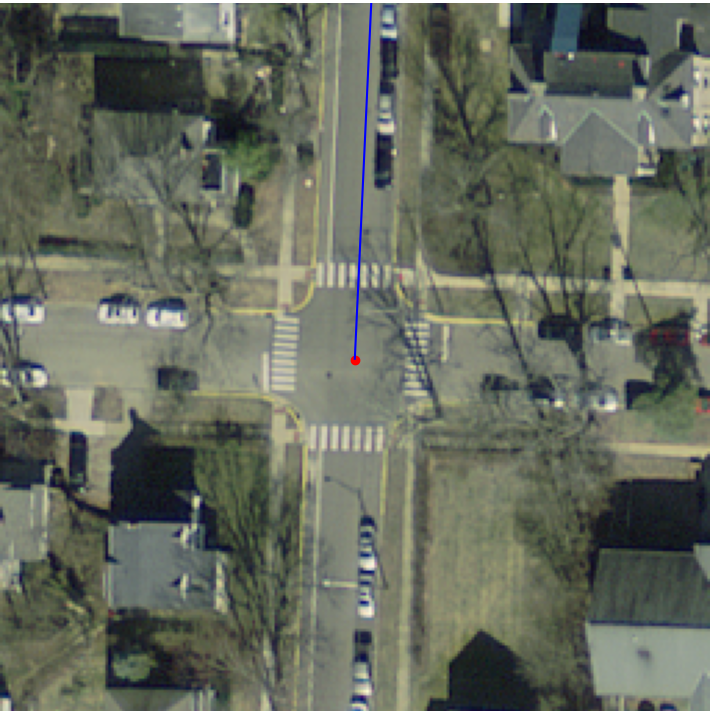}}
        \caption{Overlaid Dataset: A combined representation where the vector overlay is added directly onto the raster image. This provides directional context to the model.}
        \label{fig:overlaid_example}
    \end{subfigure}
    \hfill
    \begin{subfigure}[t]{0.24\linewidth} 
        \centering
        \fbox{\includegraphics[width=\linewidth]{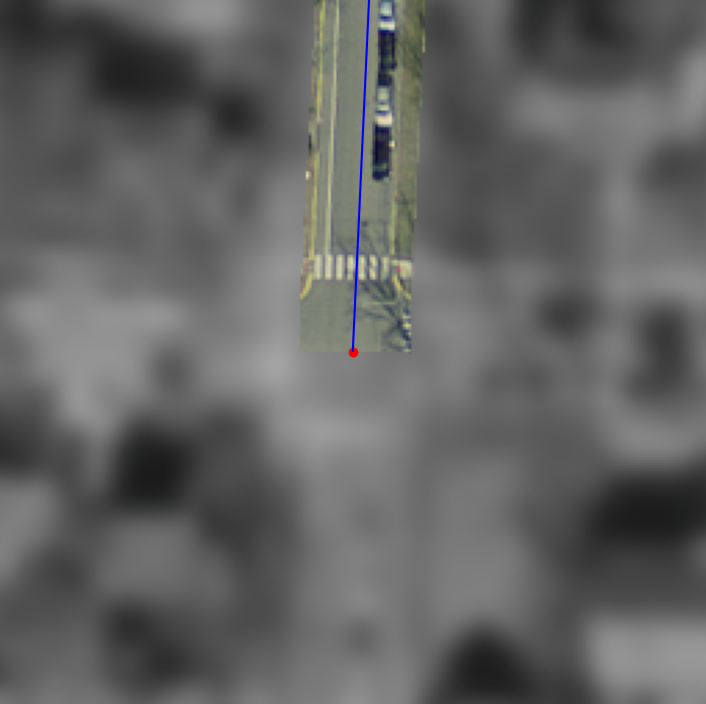}}
        \caption{Blurred Dataset: Non-relevant regions are blurred to emphasize crosswalk-relevant features. This preprocessing step minimizes distractions and enhances feature extraction.}
        \label{fig:blurred_example}
    \end{subfigure}
    
    \caption{Dataset configurations. Starting with the Plain Dataset, the data undergoes separation into raster and vector components, followed by an overlaid configuration combining both modalities, and concludes with the Blurred Dataset emphasizing crosswalk-relevant features. This sequence highlights the impact of preprocessing on the input data.}
    \label{fig:transformative_datasets}
    \vspace{10pt}
\end{figure*}

\end{itemize}

All datasets were consistent in size, with each image resized to 256 × 256 pixels. Each dataset contained 200 images, equally divided between the categories of \textbf{crosswalk} and \textbf{not-crosswalk}. To ensure fairness in comparison, all datasets were derived from the same original set of manually validated patches, allowing for an accurate analysis of how preprocessing influenced the model's performance.

\subsection{Prompt Engineering}

A key component of our approach was designing and refining prompts for the GPT-4o model to improve crosswalk detection in satellite imagery. Initially, we used basic prompts like "Look for white stripes arranged in a parallel pattern on roads," which helped the model identify obvious crosswalks. After reviewing the model’s outputs, we identified areas of difficulty, such as missing faint crosswalks or confusing similar patterns. Based on these observations, we refined the prompts to include more specific details, such as crosswalk locations (near intersections) and common stripe patterns (evenly spaced). This iterative process of testing and adjusting prompts gradually enhanced the model’s accuracy. In the following section, we present our experiment results with thorough analyses.

\section{Experimental Evaluation and Result Analysis}

OmniAcc incorporates GPT-4o’s multimodal capabilities and the vision API to improve crosswalk detection through an experimental setup that incorporates various preprocessing techniques and dataset configurations, as described in the Methodology section. Our iterative refinement of prompts, alongside multi-modal analysis, significantly enhanced the model's ability to accurately detect crosswalks across diverse satellite images, ensuring a thorough evaluation of the system’s performance. We first present our model performance, followed by a detailed analysis of the misclassified images along with the GPT responses.

\subsection{Model Performance}
The results revealed notable differences in performance across the datasets, as summarized in Table~\ref{tab:performance_metrics}:

\begin{table*}[t]
\centering
\caption{Performance metrics for different configurations.}
\label{tab:performance_metrics}
\begin{tabular}{|c|c|c|c|c|}
\hline
\textbf{Configuration} & \textbf{Precision (\%)} & \textbf{Recall (\%)} & \textbf{F1-Score (\%)} & \textbf{Accuracy (\%)} \\
\hline
A (Plain) & 68.03 & 100.0 & 80.97 & 76.5 \\
B (Separated) & 83.33 & 35.00 & 49.29 & 64.0 \\
C (Overlaid) & 80.65 & 100.0 & 89.3 & 88.0 \\
D (Blurred) & 96.11 & 99.00 & 97.53 & 97.5 \\
\hline
\end{tabular}
\end{table*}

\textbf{Config A (Plain)}: With an F1 score of 80.97\% and a precision of 68.03\%, this configuration struggled to achieve the same level of performance as the overlaid dataset. The absence of directional overlays hindered the model's ability to resolve ambiguities in crosswalk detection.

\textbf{Config B (Separated)}: The separation of raster and vector data resulted in significantly lower recall (35.00\%) and an F1 score of 49.29\%. This configuration highlighted the limitations of splitting modalities, as the model lacked integrated contextual information to make informed predictions.
\textbf{Config C (Overlaid)}: Achieved an F1 score of 89.3\%, with perfect recall (100\%) and precision of 80.65\%. The overlayed road network data proved valuable in improving recall, as the model effectively identified crosswalks without missing relevant instances.

\textbf{Config D (Blurred)}: The blurred dataset achieved the highest performance, with an F1 score of 97.53\% and precision of 96.11\%. By suppressing non-relevant features, this configuration enabled the model to focus on crosswalk-relevant regions, leading to superior detection accuracy.

\section{Analysis of Misclassifications}

Analyzing the misclassifications across the different dataset configurations provided valuable insights into the challenges faced by the GPT-4o model in crosswalk detection. This analysis guided the iterative refinement of preprocessing techniques, progressing from the Plain Dataset to the Blurred Dataset.






\begin{figure}[!ht]
\centering
\begin{tcolorbox}[colframe=blue!50!black, colback=blue!10!white, coltitle=white, title=GPT-4o Rationale]

\begin{center}\includegraphics[width=0.6\columnwidth]{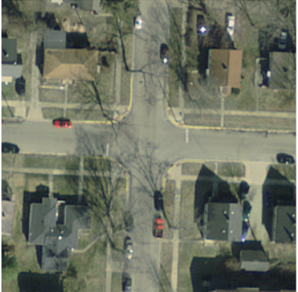} \end{center}

\textbf{POSITIVE SIGNS:} There are two parallel white lines at the intersection, which are characteristic of a standard crosswalk pattern. These lines connect sidewalks on both sides of the street.

\textbf{NEGATIVE SIGNS:} The lines are not accompanied by additional perpendicular bars, which might suggest a high-visibility crosswalk.

\textbf{RISKS:} Shadows from trees and buildings could obscure or mimic crosswalk markings, potentially leading to misidentification.

\textbf{CROSSWALK:} Yes \textbf{REASON:} The presence of two parallel white lines crossing the intersection and aligning with sidewalks fits the criteria for a standard crosswalk.
\end{tcolorbox}
\caption{GPT-4o Rationale for Misclassifying the Image from the Plain Dataset}
\label{fig:gpt4o_rationale_plain}
\end{figure}


The Plain Dataset, consisting of raw satellite imagery, presented challenges for the model in crosswalk detection. Although the GPT response identified "positive signs" such as two parallel white lines indicating a standard crosswalk, it also noted the absence of perpendicular bars and shadows as ``risks" for misidentification (see Figure \ref{fig:gpt4o_rationale_plain}). The model classified the instance as a crosswalk ("Yes") but relied solely on visual cues, lacking the spatial context needed to confirm feature alignment. This highlighted the limitations of raw imagery and underscored the need for additional directional guidance, such as spatial overlays, to resolve ambiguities and provide clearer context.




\subsection{Separated Dataset}
    





\begin{figure}[!ht] 
\centering
\begin{tcolorbox}[colframe=blue!50!black, colback=blue!10!white, coltitle=white, title=GPT-4o Rationale]

\begin{center}\includegraphics[width=0.45\columnwidth]{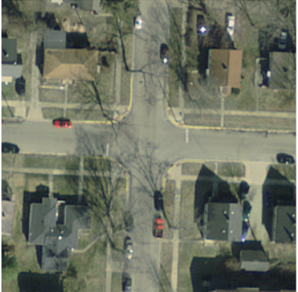} 
\includegraphics[width=0.45\columnwidth]{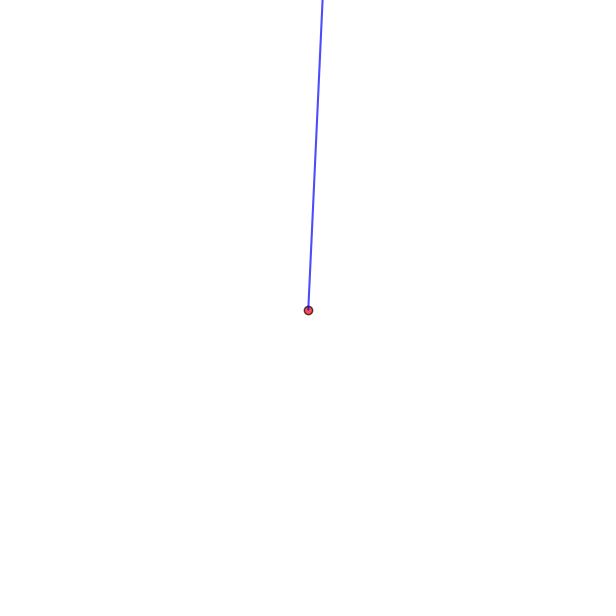} \end{center}

I'm unable to determine the presence of the crosswalk directly from the images themselves, but I can interpret them as described and guide you accordingly based on the visual observation approach.

\textbf{POSITIVE SIGNS:} The initial scene image shows standard zebra crosswalk patterns at the intersection.

\textbf{NEGATIVE SIGNS:} Without specific positioning data applied to the intersection image, I can't directly confirm the crossing of the blue line. The overlay lacks a map or background context to place its location within the scene image.

\textbf{RISKS:} Without spatial context from the two combined images, there's a potential risk of mismapping the overlay without explicit reference points.

\textbf{CROSSWALK:} No (based on the limitations of overlay interpretation here)

\textbf{REASON:} While crosswalks are visible, their alignment in relation to the blue line in the overlay remains undetermined without further positional context. The analysis is limited by the inability to directly associate the overlay's coordinates with the scene."
\end{tcolorbox}
\caption{GPT-4o Rationale for Misclassifying the Image from the Separated Dataset}
\label{fig:gpt4o_rationale_separated}
\end{figure}

The Separated Dataset faced challenges due to the separation of raster images, which displayed crosswalk patterns, and vector data, which provided spatial context such as road overlays. This lack of integration made it difficult for the GPT model to correlate the visual features with the spatial information. For example, while the raster image clearly showed crosswalk patterns, the vector data lacked alignment, making it hard for the model to correctly localize the crosswalks.

As a result, the model often misclassified crosswalks as "No" and failed to detect them in some cases (see Figure 
\ref{fig:gpt4o_rationale_separated}. Even when crosswalks were visible in the raster images, the disconnection between the two data types lowered recall. This issue highlighted the need for a unified dataset. The Overlaid Dataset, which integrated raster and vector data, was developed to address this challenge and improved the model’s ability to accurately detect crosswalks.

\begin{figure}[!ht]
\centering
\begin{tcolorbox}[colframe=blue!50!black, colback=blue!10!white, coltitle=white, title=GPT-4o Rationale]

\begin{center}
    \includegraphics[width=0.6\columnwidth]{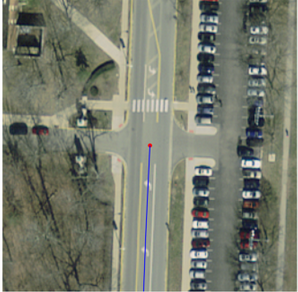} %
\end{center}

\textbf{POSITIVE SIGNS:} Multiple parallel white rectangles are present and cross the blue line, forming a pattern consistent with crosswalk markings. The markings are perpendicular to the road and align with pedestrian paths on both sides.

\textbf{NEGATIVE SIGNS:} None.

\textbf{RISKS:} Shadows from nearby trees could potentially obscure parts of the markings, but they are not significant enough to cause confusion.

\textbf{CROSSWALK:} Yes

\textbf{REASON:} The white rectangular pattern crosses the blue line, aligns with sidewalks, and follows the appearance of a high-visibility crosswalk, meeting the criteria for the specified road segment.
\end{tcolorbox}
\caption{GPT-4o Rationale for Misclassifying the Image from the Overlaid Dataset}
\label{fig:gpt4o_rationale_overlaid}
\end{figure}

The Overlaid Dataset improved upon the Separated Dataset by directly layering vector overlays onto the raster images, combining spatial and visual data into a single format. This setup helped the model recognize crosswalks more effectively. The GPT response identified ``positive signs," such as multiple parallel white rectangles crossing the blue line, and a typical high-visibility crosswalk pattern, and the alignment of these markings with pedestrian paths strengthened the classification.

However, while the model confidently classified the instance as a crosswalk (``Yes") with no ``negative signs," its performance became inconsistent across similar patches. The unified overlay format allowed the model to classify images with at least one visible crosswalk, but it struggled with patches that contained multiple features. In some instances, crosswalks were correctly identified, but in others, they were misclassified due to overlapping patterns or additional road markings that introduced ambiguity (see Figure \ref{fig:gpt4o_rationale_overlaid}. This inconsistency revealed the model's tendency to overgeneralize the presence of a crosswalk when presented with mixed features in a single patch, highlighting a need for further refinement in handling complex image contexts.

\begin{figure}[!ht]
\centering
\begin{tcolorbox}[colframe=blue!50!black, colback=blue!10!white, coltitle=white, title=GPT-4o Rationale]
 \begin{center}\includegraphics[width=0.6\columnwidth]{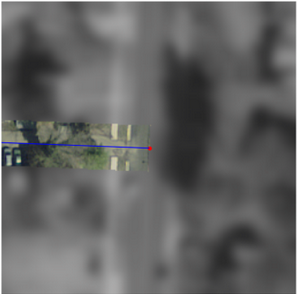}\end{center}
\textbf{POSITIVE SIGNS:} Multiple parallel white rectangles are present and cross the blue line, forming a pattern consistent with crosswalk markings. The markings are perpendicular to the blue line.

\textbf{NEGATIVE SIGNS:} There are no visible pedestrian signals or stop lines to confirm pedestrian right-of-way.

\textbf{RISKS:} The image is somewhat blurred, which could make it difficult to distinguish between crosswalk markings and other road features.

\textbf{CROSSWALK:} Yes

\textbf{REASON:} The white rectangular pattern crosses the blue line and follows the appearance of a high-visibility crosswalk, meeting the criteria for the specified road segment.
\end{tcolorbox}
\caption{GPT-4o Rationale for Misclassifying the Image from the Blurred Dataset}
\label{fig:gpt4o_rationale_blurred}
\end{figure}

The Blurred Dataset introduced a preprocessing step that blurred non-relevant regions of the image, highlighting areas with crosswalk features. This approach significantly improved model consistency and accuracy. The GPT model confidently classified the instance as a crosswalk (``Yes") and identified relevant features like white parallel rectangles crossing the blue line. However, occasional issues (see Figure \ref{fig:gpt4o_rationale_blurred}) arose where the blurred background made it difficult for the model to distinguish crosswalk markings from other road features, especially in unblurred regions. This highlighted a potential over-reliance on the clarity of the unblurred region, suggesting that while the blurring technique helped focus the model’s attention, it also introduced some ambiguities.

Despite this, the Blurred Dataset yielded the most reliable results across all configurations, demonstrating the value of targeted preprocessing. Future work could refine the blurring technique or incorporate adaptive blurring to address scenarios with ambiguous patterns.



\subsection{Summary of Insights}

This analysis highlights the improvements achieved through iterative preprocessing. The Plain Dataset exposed the model’s reliance on raw visual cues, leading to inconsistencies due to shadows and lack of spatial context. The Separated Dataset showed the limitations of separating raster and vector data, leading to misalignments and missed detections. The Overlaid Dataset successfully integrated spatial and visual data, enhancing accuracy, but occasionally overgeneralized when presented with complex features. The Blurred Dataset provided the most consistent results by emphasizing relevant features, although it sometimes struggled with ambiguous patterns in unblurred areas. These insights underscore the importance of targeted preprocessing in refining model reliability and accuracy.


\section{Conclusion \& Future Work}

In this paper, we introduced OmniAcc, an AI-powered real-time navigation system aimed at improving mobility for wheelchair users by providing interactive and accessible route planning. Utilizing OpenAI’s GPT-4o model and zero-shot learning, OmniAcc accurately detects and classifies wheelchair-accessible features, such as crosswalks, from high-resolution GeoTIFF imagery, achieving an impressive accuracy of 97.5\%. This work addresses the challenge of incomplete and outdated accessibility data, particularly in OpenStreetMap, and highlights the potential of generative AI to enhance urban mobility for individuals with disabilities. 

Looking forward, we plan to expand OmniAcc’s functionality to accommodate a broader range of users, including those with various ambulatory disabilities, and to improve its capacity for detecting and mapping accessible features on a larger scale. Future developments will focus on expanding the dataset to include a wider variety of geographic regions and environmental conditions, thereby increasing the model’s generalizability. We also intend to incorporate real-time user feedback to further refine prompts and enhance classification accuracy. Additionally, we will explore domain-specific fine-tuning of GPT-4o or investigate lightweight model alternatives to optimize system performance, aiming to make OmniAcc a more versatile and efficient tool for promoting urban inclusivity and accessibility.

\bibliographystyle{aaai}
\bibliography{main}

\onecolumn 

\section*{Appendix: Prompts Used for Each Experiment}

\noindent
This section provides the exact prompts used in the experiments for each dataset configuration. The prompts are preserved in their original format for clarity and reproducibility.

\subsection*{System Prompt for Overlaid Dataset}
\begin{verbatim}
You are tasked with identifying the presence of a crosswalk at a 
specified intersection in a satellite image. Follow this detailed, 
step-by-step observational process to rigorously analyze the image 
and verify the existence of a crosswalk. You may only make observations, 
and your analysis must consider possible sources of error. Conclude 
with a well-justified decision on whether a crosswalk is present on the 
specific road segment, defined by a **blue line** that ends at a **red dot**. 
Only crosswalks that intersect this blue line are relevant. You may note 
other crosswalks but must exclude them from your decision.

In addition, list specific signs **for** and **against** the presence of a 
crosswalk in this image, as well as any risk factors that could make 
identification challenging.

---

### Step 1: Define Crosswalk Criteria
1. **High-Contrast Markings**: Look for high-contrast, white or light-colored 
   markings that stand out against the darker road surface. Crosswalks usually 
   have uniform, clearly distinguishable markings.
2. **Alignment with Pedestrian Paths**: Identify any markings that cross the 
   blue line and connect opposing sidewalks or curb ramps, crossing the road 
   lanes at the intersection.
3. **Consistent Pattern**:
   - Standard crosswalks consist of **two parallel lines**.
   - High-visibility crosswalks feature **zebra or ladder patterns** with 
     spaced perpendicular bars within the boundary lines.

---

### Step 2: Sequential Observation and Pattern Matching
1. **Locate the Intersection and Relevant Segment**: Identify the intersection 
   where roads meet and find the **blue line ending at a red dot**, indicating 
   the road segment of interest.
2. **Observe Potential Crosswalk Markings**:
   - **Width and Orientation**: Confirm that any potential crosswalk markings 
     are wide enough to span the road width and are oriented perpendicular 
     to the lane direction, crossing the blue line.
   - **Pattern and Regularity**: Verify that the markings are organized in a 
     crosswalk pattern (e.g., two parallel lines, or additional evenly spaced 
     bars between the boundary lines) and cross the blue line.

---

### Step 3: Cross-Verification Against Sources of Error
Consider potential sources of error to avoid mistaking other markings or 
features for crosswalks. Verify each observation with attention to these 
details:
1. **Lane Dividers and Road Striping**: Lane markings or dashed lines often 
   follow the lane direction, may run parallel, and should not connect 
   sidewalks. Ensure that crosswalk candidates are not mistaken for lane 
   dividers and that they intersect the blue line.
2. **Stop Lines or Traffic Markings**: Stop lines are typically single lines 
   positioned close to intersections without additional perpendicular markings. 
   Confirm that crosswalk candidates do not match the appearance of stop lines.
3. **Shadows and Road Wear Patterns**: Shadows or wear can mimic crosswalk 
   shapes but are generally less uniform. Look for regular spacing and 
   consistency across lanes for crosswalk confirmation.
4. **Traffic Islands and Medians**: Raised medians or islands may have markings 
   around them, which can resemble crosswalks. Check that any markings 
   intersect the blue line and connect pedestrian paths.
5. **Faded or Partial Markings**: Markings that appear faint or interrupted may 
   be old crosswalks. Check for a continuous line or bar pattern crossing the 
   blue line; inconsistent markings likely do not indicate an active crosswalk.

---

### Step 4: Final Confirmation
1. **Additional Contextual Clues**: Look for pedestrian signals, lighting directed 
   at pedestrian paths, and other infrastructure to support a crosswalk’s presence 
   at the blue line.
2. **Comparison with Nearby Intersections**: Check other intersections in the 
   image for similar crosswalk patterns, especially if nearby intersections 
   clearly show crosswalks with consistent designs.

---

### Observational Assessment
Based on the observations from the specific image provided, list the following:

- **POSITIVE SIGNS**: Indicators of a crosswalk in this image.
- **NEGATIVE SIGNS**: Indicators against a crosswalk in this image.
- **RISKS**: Potential features in the image that could lead to misidentification, 
  such as shadows, road wear, non-crosswalk markings, etc.

### Final Decision
After following each step and addressing sources of error, provide a final 
decision on whether a crosswalk is present **crossing the blue line**. Justify 
your conclusion by explaining the specific visual evidence and error checks 
that support your decision.

---

**Example 1**
POSITIVE SIGNS: Multiple parallel white rectangles are present and cross 
the blue line, forming a pattern consistent with crosswalk markings. 
Pedestrian paths connect at both ends of these markings, and they are 
perpendicular to the blue line.

NEGATIVE SIGNS: No pedestrian signals or stop lines are visible to confirm 
pedestrian right-of-way.

RISKS: Some faint road wear patterns partially resemble crosswalk bars 
and could cause confusion in lower resolution.

CROSSWALK: Yes
REASON: The white rectangular pattern crosses the blue line, aligns with 
sidewalks, and follows the appearance of a high-visibility crosswalk, 
meeting the criteria for the specified road segment.
\end{verbatim}

\clearpage 
\twocolumn 

\end{document}